\documentclass[sigconf, ann]{acmart}

\usepackage{booktabs}
\usepackage{subfig}
\usepackage{multirow}
\usepackage{amsthm}
\usepackage{hyperref}
\usepackage{algorithm}
\usepackage{algorithmic}
\usepackage{bm}
\usepackage{color}
\usepackage{colortbl}
\usepackage{dutchcal}

\usepackage{mathtools}
\usepackage{xcolor}
\AtBeginDocument{%
  \providecommand\BibTeX{{%
    \normalfont B\kern-0.5em{\scshape i\kern-0.25em b}\kern-0.8em\TeX}}}

\acmSubmissionID{167}

\begin{document}

\title{MonoTDP: Twin Depth Perception for Monocular 3D Object Detection 
in Adverse Scenes}



\author{Xingyuan Li}
\email{xingyuan_lxy@163.com}
\affiliation{
  \institution{Dalian University of Technology}
  \city{Dalian}
  \state{Liaoning}
  \country{China}
}

\author{Jinyuan Liu}
\email{atlantis918@hotmail.com}
\affiliation{
  \institution{Dalian University of Technology}
  \city{Dalian}
  \state{Liaoning}
  \country{China}
}

\author{Yixin Lei}
\email{yixinlei121@gmail.com}
\affiliation{
  \institution{Dalian University of Technology}
  \city{Dalian}
  \state{Liaoning}
  \country{China}
}

\author{Long Ma}
\email{longma@mail.dlut.edu.cn}
\affiliation{
  \institution{Dalian University of Technology}
  \city{Dalian}
  \state{Liaoning}
  \country{China}
}

\author{Xin Fan}
\email{xin.fan@dlut.edu.cn}
\affiliation{
  \institution{Dalian University of Technology}
  \city{Dalian}
  \state{Liaoning}
  \country{China}
}

\author{Risheng Liu}
\email{rsliu@dlut.edu.cn}
\affiliation{
  \institution{Dalian University of Technology}
  \city{Dalian}
  \state{Liaoning}
  \country{China}
}
\renewcommand{\shortauthors}{xx} 

\begin{abstract}
	3D object detection plays a crucial role in numerous intelligent vision systems. Detection in the open world inevitably encounters various adverse scenes, such as dense fog, heavy rain, and low light conditions. Although existing efforts primarily focus on diversifying network architecture or training schemes, resulting in significant progress in 3D object detection, most of these learnable modules fail in adverse scenes, thereby hindering detection performance. To address this issue, this paper proposes a monocular 3D detection model designed to perceive twin depth in adverse scenes, termed MonoTDP, which effectively mitigates the degradation of detection performance in various harsh environments. Specifically, we first introduce an adaptive learning strategy to aid the model in handling uncontrollable weather conditions, significantly resisting degradation caused by various degrading factors. Then, to address the depth/content loss in adverse regions, we propose a novel twin depth perception module that simultaneously estimates scene and object depth, enabling the integration of scene-level features and object-level features. Additionally, we assemble a new adverse 3D object detection dataset encompassing a wide range of challenging scenes, including rainy, foggy, and low light weather conditions, with each type of scene containing 7,481 images. Experimental results demonstrate that our proposed method outperforms current state-of-the-art approaches by an average of 3.12\% in terms of $AP_{R40}$ for car category across various adverse environments.
\end{abstract}

%




\maketitle

\section{Introduction}
	In general, vision-based object detection plays a critical role in autonomous driving and infrastructure-independent robot navigation. These detection techniques are employed to interpret the surrounding environment by recognizing and categorizing object instances, as well as determining their spatial positions and orientations. Although recent advances in 2D object detection\cite{he2017mask,zhu2020deformable} have shown significant improvements in both accuracy and processing speed, 3D object detection remains a more complex task, as it strives to simultaneously determine the pose and location of each object.

    \begin{figure}[ht]
        \centering
        \includegraphics[scale=0.45]{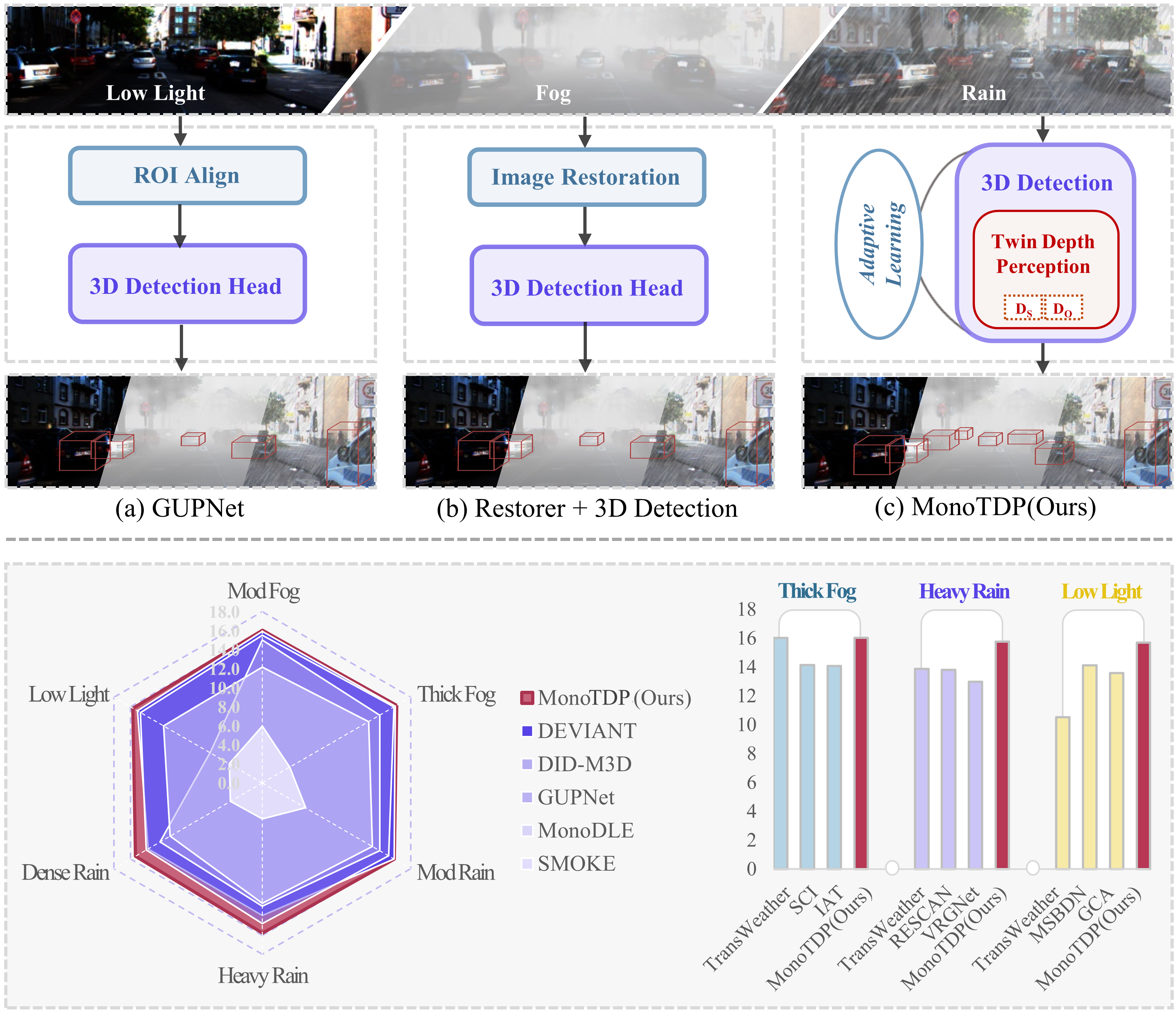}
		\caption{The top row illustrates 3D object detection methods in adverse scenes: (a) existing models overlooking environmental context, causing errors; (b) conventional solutions using image restoration, potentially yielding unsuitable images; (c) Our MonoTDP method that employs adaptive learning strategy to acclimate to harsh weather by penalizing perceptual errors and improving instance depth estimation via perceiving scene depth. The bottom row showcases the performance comparison, revealing MonoTDP's superior average precision across all six weather conditions, highlighting its significant performance enhancement.}
	\end{figure}

	A common approach to 3D object detection frequently involves the use of LiDAR sensors or stereo cameras for depth estimation\cite{zhou2018voxelnet,xu2021spg,sheng2022rethinking, yang2022graph}. Although these methods can provide accurate depth information, they significantly increase the cost of implementing practical systems. Consequently, monocular 3D object detection\cite{chen2016monocular,ma2021delving} has emerged as a promising alternative, garnering considerable interest within the research community.

	Over the past few years, numerous monocular 3D object detection techniques have been proposed and successfully implemented. These approaches can be broadly classified into two categories: those based on single images \cite{M3d-rpn,ma2021delving} and those leveraging auxiliary information \cite{manhardt2019roi,wang2019pseudo}. Single-image based techniques, such as M3D-RPN\cite{M3d-rpn} and MonoDLE\cite{ma2021delving}, primarily focus on extracting depth information from a single input image. They employ innovative strategies like depth-aware convolution and depth error analysis to enhance detection performance, offering cost-effectiveness and simplicity in the process. On the other hand, auxiliary-information based methods, including RoI-10D\cite{manhardt2019roi} and Pseudo-LiDAR\cite{wang2019pseudo}, incorporate supplementary data sources, such as CAD models or point clouds, to enrich the detection process. This additional information helps to overcome the limitations inherent in single-image based approaches, leading to more robust and accurate object localization and classification. Collectively, these methodologies have significantly advanced the field of 3D object detection, enabling more accurate and reliable object localization and classification in various application scenarios.
	
	Nevertheless, these methods have encountered several issues, primarily in the following three aspects. $(\romannumeral1)$ 3D object detection is inevitable to face real-world adverse conditions, such as rain occlusions, fog-induced scattering, and texture loss in low-light situations. These conditions lead to degraded image quality, partial or complete occlusion of objects, blurring effects, and reduced contrast, ultimately undermining the detection performance. $(\romannumeral2)$ Monocular 3D detection is inherently limited by the single viewpoint, making it difficult to recover depth information from a single 2D image, as the camera projection process results in the loss of spatial information. This limitation can lead to ambiguities and uncertainties in depth estimation, posing challenges for accurate object localization and classification. $(\romannumeral3)$ The scarcity of datasets hinders advancements in the field. There is a lack of comprehensive datasets capturing a wide range of adverse weather conditions and complex environments characteristic of real-world driving scenarios. This deficiency restricts the learning of complementary information and validation of detection algorithm effectiveness, thus impeding progress in developing more robust and adaptable 3D object detection techniques.

	To address the aforementioned issues, we propose a novel monocular 3D object detection method tailored for challenging environments, dubbed MonoTDP. Our approach incorporates an adaptive learning strategy and a twin depth perception module. The adaptive learning strategy, during training, penalizes incorrect perception of harsh conditions, fostering robust multi-environment capabilities. This allows our method to better adapt to complex scenarios, such as rain, fog, and low light. The twin depth perception module tackles depth ambiguity by estimating both scene and object depth using scene-level and object-level features, effectively recovering missing depth cues in degraded regions. We introduce a diverse dataset covering various challenging scenarios, including moderate fog, thick fog, dense fog, moderate rain, heavy rain, dense rain, and low light conditions. Each category consists of 7,481 images. Figure 1 demonstrates that our proposed model outperforms state-of-the-art(SOTA) 3D object detectors and cascade of image enhancement and 3D detection models. Our contributions are four-fold:

	\vspace{-1em}
    \begin{itemize}
	\item We introduce a robust network specifically designed to handle a variety of adverse environments, significantly improving the performance and resilience of monocular 3D object detection models across a wide range of challenging real-world situation.
	\item We propose an adaptive learning strategy implemented during the training process, enabling the model to acclimatize to inclement weather conditions and augment the precision of 3D object detection. This component aids the model in discerning various environments and extracting resilient features that remain less susceptible to degrading factors.
	\item A twin depth perception module is incorporated to address the challenge of depth estimation in adverse settings by concurrently leveraging object-level features for object depth estimation and scene-level features for scene depth estimation. This approach mitigates the problem of information and depth loss under diverse unfavorable conditions.
	\item To support 3D object detection in harsh environments, we have compiled a comprehensive dataset comprising 7,481 images for each of the seven demanding conditions: moderate fog, thick fog, dense fog, moderate rain, heavy rain, torrential rain, and low light.
	\end{itemize}

\begin{figure*}[ht]
    \centering
    \includegraphics[scale=0.56]{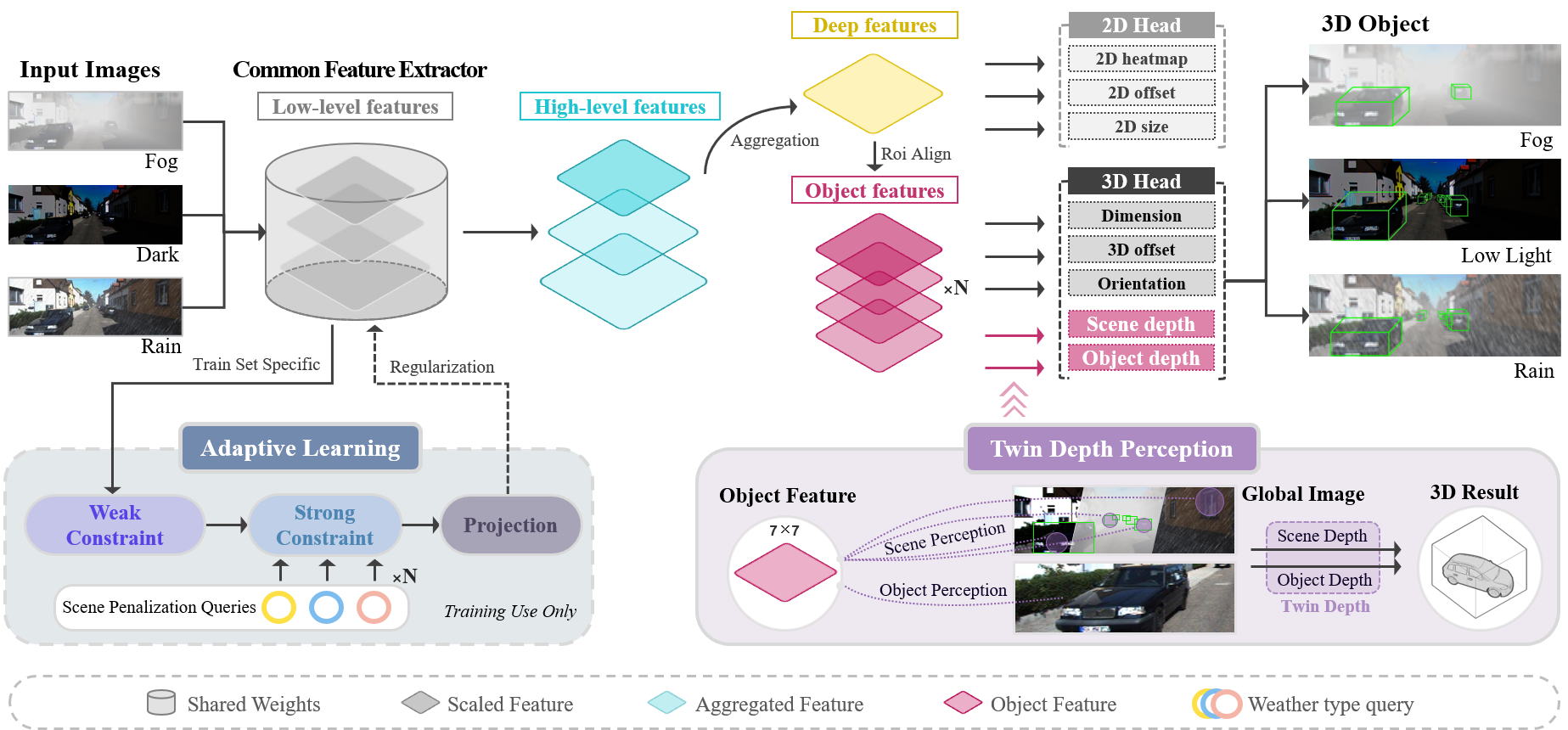}
	\caption{The pipeline of MonoTDP.  It consists of two core components: Adaptive Learning (Sec. 3.2) regularizes features from Common Feature Extractor to help model perceive clean meta features that are not degraded by adverse factors, and it is only used in training stage. Twin depth perception module (Sec. 3.3) is conducted in 3D object detection under adverse weather conditions to obtain better instance depth by combining scene depth and object depth.}
\end{figure*}

\section{Related Work}
\subsection{Monocular 3D Object Detection}
Monocular 3D object detection can be roughly divided into single-image based methods and auxiliary-information based methods.

The most commonly used methods take only one image as input and output the 3D information of the object in the image. To estimate the depth, M3D-RPN\cite{M3d-rpn} designed a depth-aware convolution that can better obtain 3D area proposals to perceive depth information. For the simplicity and effectiveness of the model, SMOKE\cite{liu2020smoke} and FCOS3D\cite{wang2021fcos3d} proposed a one-stage monocular 3D detection model based on CenterNet\cite{duan2019centernet} and FCOS\cite{tian2019fcos} respectively. MonoDLE\cite{ma2021delving} obtained more reliable results by analyzing the manually designed depth error. MonoFLEX\cite{zhang2021objects} used uncertainty-guided depth and adopted special treatments for different objectives.

Plenty of approaches use additional data to assist the learning of monocular 3D detection models. RoI-10D\cite{manhardt2019roi} used CAD models to introduce prior knowledge to enhance training samples. Pseudo-LiDAR\cite{wang2019pseudo} proposed to lift the estimated depth to the point cloud and then use a detector based on the LiDAR method. DID-M3D\cite{peng2022did} used a dense depth map. However, the previous works do not consider the impact of complex environments. Thus, we design a model that can adapt to various adverse scenes.

\subsection{Degradation Factor Removal}
It has been widely explored to remove degradation factors from images in adverse weathers, such as rain removal\cite{li2020all,wang2020model,yang2019joint,liu2018deep,liu2019knowledge},fog removal\cite{liu2020learning,cai2016dehazenet,li2018single,zhang2018density,liu2018learning}, low-light enhancement\cite{liu2020bilevel,liu2019compounded,liu2022target,zhang2019kindling,zamir2020learning,chen2018learning,liu2021learning}.

To remove rain, \cite{yang2019joint} split the rain streak into different layers, \cite{zhang2019image} used a GAN-based method and \cite{quan2019deep} used a dual attention mechanism. To cope with haze, DehazeNet\cite{cai2016dehazenet} restored the visibility of images through a scattering transformation. DCPDN\cite{zhang2018densely} generated transmission map, atmospheric light and dehazing map. \cite{zhang2021hierarchical} designed a hierarchical dense perceptual network. To enhance low-light, LLNet\cite{lore2017llnet} adaptively brightened images through multi-layer encoders. \cite{shen2017msr,tao2017llcnn,lv2018mbllen,ren2019low} used multi-scale features to better restore clear images. \cite{shen2017msr} used two deep networks to decouple images.

Furthermore, there are some all-in-one\cite{li2020all} networks. \cite{li2020all} used multiple discriminative encoders to deal with different environments. U-former\cite{wang2019spatial}, Swin-IR\cite{liang2021swinir} proposed to solve image restoration problems in adverse environments based on different Transformers. Different from them, the method we proposed can both handle multiple adverse scenes and benefit 3D object detection through an adaptive learning strategy.

\subsection{Depth Estimation}
There are also methods that utilize geometric constraints and scene priors to use auxiliary data to help depth estimation. The early work \cite{mousavian20173d} used 2D-3D box geometric constraints to estimate instance depth. But this indirect approach which does not fully use of supervisions has poor performance. \cite{li2020rtm3d} predicted nine perspective key-points of a 3D bounding box in the image space and optimized the initially estimated instance depth by minimizing projection errors. \cite{li2021monocular} followed this line and integrated this optimization into an end-to-end training process. Recently, \cite {zhang2021objects} predicted the nine-perspective key-points and produced new instance depths by using the projection heights in pair key-points and geometric relationships.

In this paper, we propose a 3D object detection model facing adverse scenes that can benefit from twin depth perception and adaptive learning strategy.

\section{Method}
Under various inclement weather conditions, the specific spectral interaction between captured objects and the camera may be affected by the absorption and scattering caused by suspended water droplets, dust, and other particulates, resulting in the loss of depth information for 3D object detection. To address this challenge, we present a monocular 3D object detection model, namely MonoTDP, for adverse environments. As depicted in Figure 2, images undergo processing via a shared feature extractor, which constitutes a part of the 2D detection backbone and is guided by the adaptive learning strategy which effectively mitigates the interference caused by various degrading factors. Consequently, we obtain deep features, 2D bounding boxes, and fundamental 3D bounding box information, such as dimensions, 3D projected centers, and angles. Subsequently, the twin depth perception module simultaneously predicts scene depth and object depth. The integration of scene depth and object depth yields accurate inferred depth values under different degrees of challenging conditions, facilitated by the comprehensive interaction between scene-level and object-level features.

\subsection{Adverse Condition Datasets Generation}
Currently, there is a significant shortage of reliable 3D object detection datasets specifically designed for adverse weather conditions. To address this issue, we have compiled datasets encompassing a wide range of adverse weather conditions, such as fog, rain, and low light, to conduct comprehensive experiments using our proposed approach, MonoTDP.

In the literature, various weather phenomena are modeled differently based on their underlying physical properties. The process of synthesizing adverse weather conditions is primarily based on the simulation of their corresponding atmospheric effects. According to the atmospheric light attenuation theory \cite{zhang2017hazerd}, the fog condition is modeled as:

\begin{equation}
\mathbf{I}=\mathbf{B} \odot \mathbf{T} + \mathbf{A}\odot (1-\mathbf{T}),
\end{equation}

\noindent where $\mathbf{I}$ represents the degraded image, $\mathbf{B}$ denotes the background, $\mathbf{A}$ refers to the atmospheric light in the scene, and $\mathbf{T}$ signifies the light propagation formula, which can be expressed as:

\begin{equation}
\mathbf{T}=e^{-\mathbf{\beta} \mathbf{d}},
\end{equation}

\noindent where $\mathbf{d}$ corresponds to the depth value of the image, and $\beta$ is a variable to modulate the scattering effect. This model enables the accurate representation of foggy conditions by simulating the scattering of light particles due to the presence of fog.

Rain with rain streaks and fog effect \cite{li2019heavy} is modeled as:

\begin{equation}
\mathbf{I}= \mathbf{T} \odot\left(\mathbf{B}+\sum_i^n \mathbf{R}_i\right)+(1-\mathbf{T}) \odot A,
\end{equation}

\noindent where, $\mathbf{R}$ represents the raindrop residual. This model incorporates the dynamics of raindrops and their impact on the image quality, taking into account the distortion caused by rain streaks and the interaction of raindrops with the scene's atmospheric light.

In addition, the image brightness is reduced using the $\gamma$ correction method to simulate low light conditions as follows:

\begin{equation}
\mathbf{I} = \mathcal{F}(\mathbf{B}, \gamma),
\end{equation}

\noindent where $\mathcal{F}$ stands for the replacement of the look-up table, and $\gamma$ indicates the gamma value for luminance correction. This method allows us to create a more realistic representation of images captured in low light scenarios by adjusting the overall brightness and contrast of the scene.

By synthesizing these diverse adverse weather conditions in our dataset, we can effectively evaluate the performance of our proposed MonoTDP model under various challenging scenarios. The data collected for our experiments is illustrated in Figure 3.

\begin{figure}[t]
    \centering
    \includegraphics[scale=0.5]{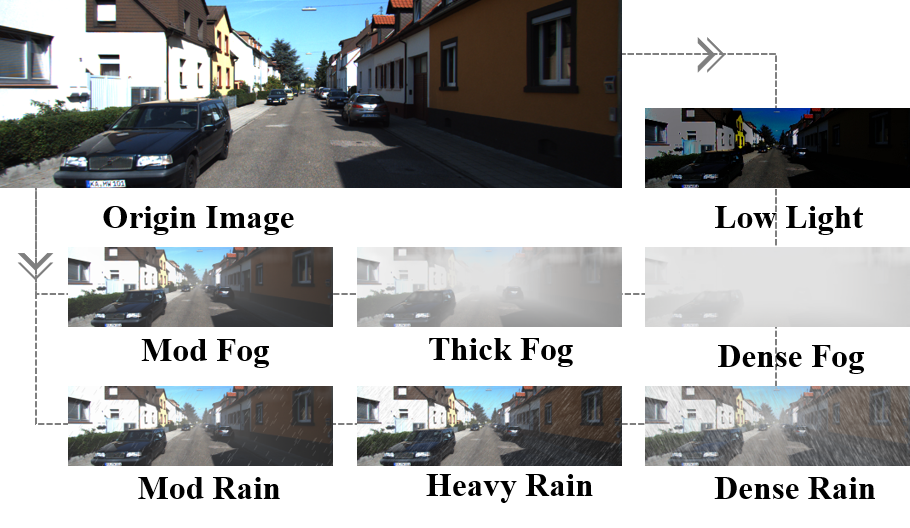}
	\vspace{-0.8em}
    \caption{Images under adverse weather conditions, conclude mod fog, thick fog, dense fog, mod rain, heavy rain, dense rain, low light.}
\end{figure}


\subsection{Adaptive Learning Strategy}
In this work, we propose a novel adaptive learning strategy comprising a weak constraint encoder and a strong constraint decoder, which are specifically designed to act as a constraint, rather than focusing on image restoration. The primary objective of this module is to facilitate the model's ability to learn and perceive intrinsic features under various adverse conditions.

The weak constraint encoder is devised to understand image features and discern the intricate characteristics of distinct adverse environments by concurrently analyzing intra-patch and normal patch features. This approach assists the model in rectifying inaccurate feature perception under a range of adverse conditions.
The strong constraint decoder employs learnable scene penalization queries to detect and penalize the model for incorrect perception across diverse environments. By doing so, it enforces the model to capture the essence of features while suppressing potential errors. The output features from both the encoder and decoder are subsequently fed to the Projection module, which serves to enhance the scene adaptability of MonoTDP. Notably, this learning strategy is only required during the training phase, thus ensuring efficiency and effectiveness in learning essential features in adverse environments.

Given a degraded image $I$ of size ${H} \times {W} \times {3} $, a common feature extractor is applied to generate low-level features ($\frac{H}{4} \times \frac{W}{4} \times C $). These features are then input into the weak constraint encoder containing SwinBlocks at different stages. We use intra-patch at each stage, where the resolution is reduced to assist the module in learning both coarse and fine contents. Figure 2 provides an overview of the adaptive learning strategy.

The weak constraint encoder is designed to extract multi-level features, thereby generating a hierarchical representation of the input image. During each stage, patch merging is utilized to decrease the resolution, and the merged features are passed on to the subsequent stage. SwinBlocks are then employed to perform feature transformation while maintaining the resolution. A SwinBlock comprises a shifted window-based $\mathrm{MSA_{SW}}$ and an MLP, as depicted in Figure 4. Layer Normalization ($(\mathrm{LN}$) is applied prior to each $\mathrm{MSA_{W}}$ and MLP module, and a residual connection is incorporated following each module. The specific computation process for two consecutive blocks is as follows:

\vspace{-0.37em}
\begin{align}
    \begin{split}
    & \bm{\hat{z}}^l = \sum_{i=1}^{h} \mathcal{w}_{i}\mathrm{MSA_{W}}\left(\mathrm{LN}\left(\bm{z}^{l-1}\right)\right) + \bm{z}^{l-1},\\
    & \bm{z}^l = \mathcal{mlp}\left(\mathrm{LN}\left(\bm{\hat{z}}^l\right)\right) + \bm{\hat{z}}^l,\\
    & \bm{\hat{z}}^{l+1} = \sum_{i=1}^{h} \mathcal{w}_{i}\mathrm{MSA_{SW}}\left(\mathrm{LN}\left(\bm{z}^l\right)\right) + \bm{z}^l,\\
    & \bm{z}^{l+1} = \mathcal{mlp}\left(\mathrm{LN}\left(\bm{\hat{z}}^{l+1}\right)\right) + \bm{\hat{z}}^{l+1},
    \end{split}
\end{align}

\noindent where $\hat{\mathbf{z}}^l$ and $\mathbf{z}^l$ represent the output of $l$th $\mathrm{MSA_{(S)W}}$ and $\mathrm{MLP}$ respectively. $\mathrm{MSA_{W}}$ and $\mathrm{MSA_{SW}}$ denote window based self-attention using traditional and shifted window. The shifted window introduces the connection between diverse parts of the feature map and the computational complexity is only linearly related to the image size. At the same time, intra-patch utilizes a similar SwinBlock as afore-mentioned. Following \cite{raffel2020exploring}, self-attention is calculated as:

\begin{equation}
    \mathbf{Attention}(Q, K, V)=\operatorname{\mathbf{SoftMax}}\left(Q K^T / \sqrt{d}+B\right) V,
\end{equation}

\noindent where $Q$, $K$, $V$ are queries keys and values that have same dimensions. $B$ is relative position bias.

In the strong constraint decoder, scene penalization queries are utilized to output a task feature vector, focusing on the multi-level features from the encoder. The decoder has only one stage but contains multiple blocks. Cross-attention is applied in this module, with K and V taken from the same output features as the last stage of the encoder, and Q being the learnable queries.

The output features of the decoder serve as the weather type task vector and are fused with the features produced by each stage of the encoder. Both output features from the encoder and decoder are then fed to the projection module and are constrained by a $smooth \, \mathcal{L_1} \, loss$. By incorporating weak constraint encoders and strong constraint decoders, the adaptive learning strategy can effectively adapt to diverse adverse environments and improve the precision of 3D object detection in adverse weather conditions. The effectiveness of this learning strategy will be further demonstrated in the subsequent ablation experiments.
\begin{figure}[t]
    \centering
    \includegraphics[scale=0.50]{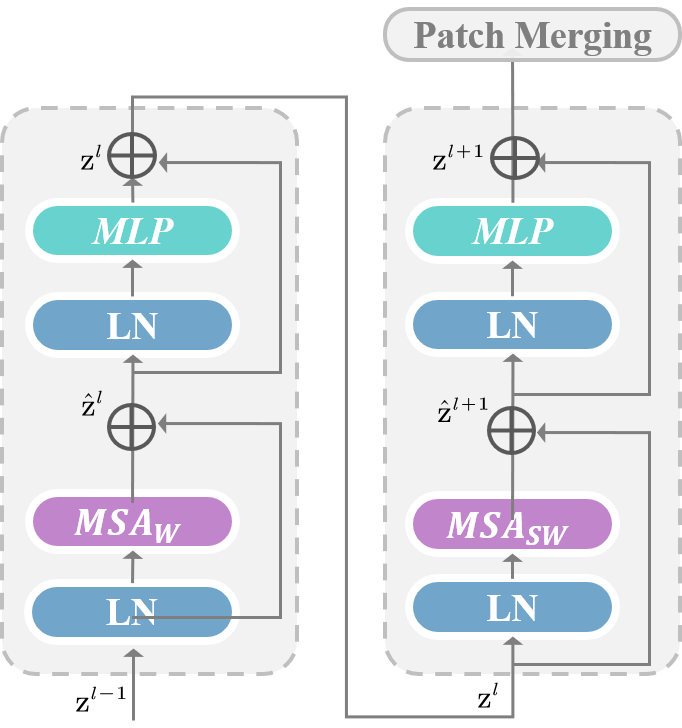}
	\vspace{-0.8em}
    \caption{The structure of SwinBlock. It contains two successive Swin Transformer Blocks (notation presented with Eq.(5)). $\mathbf{MSA_{W}}$ and $\mathbf{MSA_{SW}}$ are multi-head self attention modules with regular and shifted windowing configurations, respectively. At each stage, we perform a patch merging after SwinBlock except for the last stage.}
\end{figure}

\subsection{3D Object Detection in Adverse Scenes}
Figure 2 shows the framework of our approach, monocular 3D object detection takes an RGB image as input and constructs a 3D bounding box for the object in 3D space. The 3D bounding box consists of the object's three-dimensional center position $(x, y, z)$, size $(h, w, l)$ and direction $\Theta$ that usually refers to the observation angle.

Our monocular 3D detection network obtains the constraint features that are optimized with adaptive learning strategy which are resistant to degrading factors. These features are then forwarded to predict 2D bounding boxes. Concretely, 2D detection backbone is applied to produce high-level deep features, and then these features are aggregated to get deep features with resolution $\mathbf{F}_{} \in \mathbf{R}^{\frac{H}{8} \times \frac{W}{8} \times C}$. Subsequently, we apply three 2D detection heads in deep features $\mathbf{F}$ to predict 2D heatmap $\mathbf{H}$ which has the same size with $\mathbf{F}$, offset ${O}_{2D}$, dimension $\mathcal{S}_{2D}$. Afterwards, through using $ROI_{Align}$ in deep feature map with 2D box information, the features of the object are generated whose size is $7 \times 7$. Finally, these features are used in the 3D detection heads to predict the object 3D center offset $O_{3 d}$,  3D size $S_{3 d}$ and direction $\Theta$.

\noindent \textbf{Twin Depth Perception}. In prevailing 3D object detection approaches, the environmental context encompassing the object is frequently overlooked. As a result, instance depth is predominantly estimated based on the features proximal to the object. This approach leaves depth estimation vulnerable to the adverse effects of various weather conditions, such as the diminished visibility caused by nighttime darkness, the obfuscation induced by fog, and the visual disturbances introduced by rain. These factors can compromise the accuracy of depth estimation.

To mitigate these challenges, we introduce a novel twin depth perception module that concurrently predicts scene depth and object depth. Contrasting with traditional 3D detection heads that directly regress depth, our twin depth perception module fosters interaction between scenes and objects, transcending the limitations of single-object perception. This holistic approach empowers the model to generate more precise inferred depth values.

To clarify, we initially posit that instance depth is composed of two distinct elements: object depth and scene depth. The object depth is obtained by regressing the features of the $7\times7$ cells after performing RoI Align. Meanwhile, the scene depth is derived from a cross-attention mechanism that considers both the local features of the object and the global feature map, ensuring an accurate depth prediction even under challenging environmental conditions.

For monocular systems, scene depth is heavily contingent upon information from the entire scene, rendering it particularly sensitive to weather condition perception. Estimating scene depth generally demands a comprehensive scene perception. By integrating both scene depth and object depth predictions, our twin depth perception module substantially bolsters the model's capacity for accurate depth estimation under a wide array of challenging weather conditions, resulting in a more robust and reliable 3D object detection system.

\noindent \textbf{Depth Geometry Uncertainty}. Following \cite{peng2022did}, depth prediction is assumed to be Laplace distribution. For each object depth $d_{obj}$ and the corresponding uncertainty $u_{obj}$, they follow the Laplace distribution $La(d_{obj}, u_{obj})$. Parallel, the scene depth follows $La(d_{sce}, u_{sce})$, where $d_{sce}$ and $ u_{sce}$ denote scene depth and its uncertainty. Accordingly, the final instance depth distribution is $La(d_{ins}, u_{ins})$, where $ d_{ins}=d_{obj}+d_{sce}$ and ${u}_{ins}=\sqrt{u_{obj}^2+u_{sce}^2}$.

\subsection{Loss Functions}
Our proposed model is designed to effectively carry out 3D object detection in intricate environments. Throughout the training process, we concurrently compute the losses associated with the adaptive learning strategy and the 3D object detection task.

To accurately capture feature representations under adverse weather conditions, our adaptive learning strategy employs the $smooth_{L1}$ loss to penalize the incorrect perception of features in challenging scenarios. This loss function is formulated as follows:

\begin{equation}
    \mathcal{L}_{s m o o t h \, L_1}= \begin{cases}0.5 \mathbf{E}^2 & \text { if }|\mathbf{E}|<1 \\ |\mathbf{E}|-0.5 & \text { otherwise },\end{cases}
\end{equation}

\noindent where $\mathbf{E}$ represents the difference between the perceived scene and real scene.

For 3D object detection, the loss function is as the following formula. It can be divided into 2D detection part and 3D detection part. As shown in Figure 2, we use the 2D heatmap $H$ to indicate the rough object center on the image. Its size is $\frac{H}{8} \times \frac{W}{8} \times B$, and $B$ is the number of categories. The 2D offset ${O}_{2D}$ refers to the residual towards rough 2D centers, and $\mathcal{S}_{2D}$ denotes the 2D box height and width. We follow \cite{duan2019centernet} to use loss functions $\mathcal{L}_H$, $\mathcal{L}_{O_{2 d}}$, $\mathcal{L}_{S_{2 d}}$.

\begin{table*}[!tbh]
	\caption{Comparison of the latest 3D object detection methods on the moderate fog, thick fog, moderate rain, heavy rain, dense rain and low light dataset based on $\mathrm{AP}_{3 D}$ of the car category. All methods have been retrained on the respective environmental datasets. \textcolor{red}{Red} and \textcolor{blue}{Blue} correspond to the first and second best results, respectively. Quantitative results substantiate that our method achieves state-of-the-art performance.}
	\vspace{-0.2em}
	\centering
	\small
	\renewcommand{\arraystretch}{1.15}
	\setlength{\tabcolsep}{0.6mm}{
      \begin{tabular}{l|c|rrr|rrr|rrr|rrr|rrr|rrr}
      \toprule[1.5pt]
      \multirow{2}{*}{Methods} & \multirow{2}{*}{Venue} & \multicolumn{3}{c|}{Mod. Fog} & \multicolumn{3}{c|}{Thick Fog} & \multicolumn{3}{c|}{Mod. Rain} & \multicolumn{3}{c|}{Heavy Rain} & \multicolumn{3}{c|}{Dense Rain} & \multicolumn{3}{c}{Low Light} \\
      \cline{3-20}          &       & \multicolumn{1}{c}{\cellcolor{gray!20} Easy} & \multicolumn{1}{c}{\cellcolor{gray!20} Mod.} & \multicolumn{1}{c|}{\cellcolor{gray!20} Hard} & \multicolumn{1}{c}{\cellcolor{gray!20} Easy} & \multicolumn{1}{c}{\cellcolor{gray!20} Mod.} & \multicolumn{1}{c|}{\cellcolor{gray!20} Hard} & \multicolumn{1}{c}{\cellcolor{gray!20} Easy} & \multicolumn{1}{c}{\cellcolor{gray!20} Mod.} & \multicolumn{1}{c|}{\cellcolor{gray!20} Hard} & \multicolumn{1}{c}{\cellcolor{gray!20} Easy} & \multicolumn{1}{c}{\cellcolor{gray!20} Mod.} & \multicolumn{1}{c|}{\cellcolor{gray!20} Hard} & \multicolumn{1}{c}{\cellcolor{gray!20} Easy} & \multicolumn{1}{c}{\cellcolor{gray!20} Mod.} & \multicolumn{1}{c|}{\cellcolor{gray!20} Hard} & \multicolumn{1}{c}{\cellcolor{gray!20} Easy} & \multicolumn{1}{c}{\cellcolor{gray!20} Mod.} & \multicolumn{1}{c}{\cellcolor{gray!20} Hard} \\
      \hline
      \hline
	  SMOKE & \textit{CVPR20} & 8.86  & 5.98  & 4.53  & 5.10  & 3.31  & 2.28  & 7.33  & 5.24  & 4.03  & 5.97  & 3.78  & 2.77  & 5.64  & 3.88  & 3.21  & 5.48  & 4.03  & 3.49  \\ 
	  MonoFLEX & \textit{CVPR21} & 19.97  & 14.11  & 11.86  & 18.37  & 13.28  & 10.57   & 17.21  & 12.94  & 11.55  & 16.99  & 11.83  & 10.12  & 15.35  & 12.14  & 10.38  & 10.43  & 8.32  & 7.75  \\ 
	  MonoDLE & \textit{CVPR21} & 14.77  & 12.15  & 10.02  & 17.35  & 12.89  & 11.27  & 15.65  & 13.34  & 12.33  & 15.64  & 12.63  & 11.13  & 14.94  & 11.20  & 9.78  & 14.69  & 11.99  & 10.60  \\
      GUPNet & \textit{ICCV21} & 21.06  & 15.02  & 12.34  & 19.91  & 14.24  & 11.57   & 19.69  & 14.24  & 12.36  & 17.36  & 12.95  & 10.76  & 16.71  & 12.40  & 10.64  & 9.84  & 6.36  & 5.09  \\
      DID-M3D & \textit{ECCV22} & {\textcolor{blue}{22.75}} & 15.52  & 12.61  & 22.19  & 15.96  & 12.86  & {\textcolor{blue}{22.42}} & 15.30  & 12.43  & {\textcolor{blue}{21.40}} & {\textcolor{blue}{14.79}} & {\textcolor{blue}{12.05}} & {\textcolor{blue}{20.56}} & {\textcolor{blue}{14.07}} & 11.88  & 21.92  & 14.79  & 12.10  \\
      DEVIANT & \textit{ECCV22} & 22.74  & {\textcolor{blue}{15.92}} & {\textcolor{blue}{13.16}} & {\textcolor{blue}{22.90}} & {\textcolor{blue}{16.11}} & {\textcolor{blue}{13.25}} & 22.35  & {\textcolor{blue}{15.99}} & {\textcolor{blue}{12.45}} & 20.18  & 13.93  & 11.96  & 20.20  & 13.85  & {\textcolor{blue}{12.26}} & {\textcolor{blue}{22.40}} & {\textcolor{blue}{15.16}} & {\textcolor{blue}{12.33}} \\
      HomoLoss & \textit{CVPR22} & 14.31  & 12.27  & 11.12  & 19.32  & 13.26  & 11.51  & 18.23  & 13.19  & 12.56  & 17.69  & 13.01  & 12.23  & 16.33  & 13.40  & 10.76  & 15.88  & 13.89  & 11.42  \\ 
	  CubeR-CNN & \textit{CVPR23} & 21.11  & 14.97  & 12.55  & 20.81  & 14.77  & 12.12   & 20.37  & 14.14  & 12.38  & 22.36  & 13.67  & 11.11  & 19.17  & 13.54  & 10.99  & 20.11  & 14.37  & 11.89  \\  
	  \hline
	  \textbf{MonoTDP} & \multirow{2}[1]{*}{\textit{-}} & {\color[HTML]{FF0000} 23.13} & {\color[HTML]{FF0000} 16.03} & {\textcolor{red}{13.19}} & {\color[HTML]{FF0000} 23.24} & {\color[HTML]{FF0000} 16.28} & {\color[HTML]{FF0000} 13.35} & {\color[HTML]{FF0000} 23.08} & {\color[HTML]{FF0000} 16.01} & {\color[HTML]{FF0000} 12.98} & {\color[HTML]{FF0000} 23.06} & {\color[HTML]{FF0000} 15.77} & {\color[HTML]{FF0000} 12.92} & {\color[HTML]{FF0000} 21.31} & {\color[HTML]{FF0000} 15.40} & {\color[HTML]{FF0000} 12.52} & {\color[HTML]{FF0000} 22.55} & {\color[HTML]{FF0000} 15.70} & {\color[HTML]{FF0000} 12.80} \\
      Improvement &       & \textbf{+0.38} & \textbf{+0.11} & \textbf{+0.03} & \textbf{+0.34} & \textbf{+0.17} & \textbf{+0.10} & \textbf{+0.66} & \textbf{+0.02} & \textbf{+0.53} & \textbf{+1.66} & \textbf{+0.98} & \textbf{+0.87} & \textbf{+0.75} & \textbf{+1.33} & \textbf{+0.26} & \textbf{+0.15} & \textbf{+0.54} & \textbf{+0.49} \\
      \bottomrule[1.5pt]
      \end{tabular}
      }
\end{table*}

For the dimensions of the 3D object, we use the typically designed $\mathcal{L}_{S_{3 d}}$ and multi-bin to calculate $\mathcal{L}_{\Theta}$ for the prediction of the object observation angle. Furthermore, the position of object is recovered by using the 3D center projection and instance depth. It is achieved by predicting 3D projection offset to the 2D center, and uses smooth L1 loss function $\mathcal{L}_{O_{3 d}}$. In addition, the instance depth is decoupled into scene depth and object depth. Like \cite{peng2022did}, the depth projected by LiDAR is used as the supervision of the scene depth, and the subtraction between the instance depth and the scene depth is used as the supervision of the object depth. The Instance depth is supervised as the sum of scene depth and object depth. The instance depth loss is $\mathcal{L}_{D_{ins}}$ and uncertainty regression loss \cite{lu2021geometry} is applied as:

\begin{equation}
	\mathcal{L}_{D_{ins}}=\frac{\sqrt{2}}{u_{vins}}\left\|d_{ins}-d_{ins}^{gt}\right\|+\log \left(u_{ins}\right),
\end{equation}
	
\noindent where $u_{ins}$ denotes the uncertainty and $gt$ is the corresponding label. We set the weight of each loss term to 1.0. The overall loss is:

\begin{equation}
    \mathcal{L}=\mathcal{L}_H+\mathcal{L}_{O_{2 d}}+\mathcal{L}_{S_{2 d}}+\mathcal{L}_{S_{3 d}}+\mathcal{L}_{\Theta}+\mathcal{L}_{O_{3 d}}+\mathcal{L}_{D_{ins}}.
\end{equation}

\section{Experiments}
This section compares the results of our method for 3D object detection in various adverse environments. The experimental datasets are based on KITTI synthetic datasets of rain, fog, and low-light scenes. In which, we measure our model by two sets of comparisons. The first set with the latest monocular 3D object detection model. The second set is compared with the latest defogging model, rain removal model and low-light enhancement model. Our experiments are performed on 4 Nvidia TITAN XP.

\subsection{Datasets and Metric}
We evaluate the performance of our methods and 5 state-of-the-art under synthetic KITTI 3D dataset, comprising 7,481 images for each of moderate fog, thick fog, dense fog, moderate rain, heavy rain, dense rain, and low light conditions. as in \cite{ge2012are}. Following the methodology of \cite{chen2016monocular}, the dataset is partitioned into 3,712 sub-training sets and 3,769 validation sets. Detection outcomes are presented in three levels of difficulty, namely easy, moderate, and hard, with the moderate scores generally utilized for ranking purposes. To assess our performance, we use average precision as the evaluation metric. The 3D bounding box is represented as $\mathrm{AP3D}_{R40}$, where R40 signifies 40 recall positions. For the three aforementioned categories, the Intersection over Union (IoU) threshold for cars is set to 0.7. 

\subsection{Implementation Details}
We conduct our experiments using 4 NVIDIA RTX TITAN XP GPUs and a batch size of 8. Our implementation is built upon the PyTorch framework. We train the network for 140 epochs, following the Hierarchical Task Learning (HTL) strategy \cite{lu2021geometry}. The Adam optimizer is employed with an initial learning rate of 1e-5. We apply a linear warm-up strategy to raise the learning rate to 1e-3 during the initial 5 epochs. Subsequently, the learning rate decays at epochs 50 and 80 with a decay rate of 0.1. For the multi-bin orientation $\theta$, we set k to 12. The backbone and head architecture are designed in accordance with \cite{lu2021geometry}. Input images are resized to a resolution of 1280 × 384, with pixel values in the range of [0, 255]. The pixel intensities are then adjusted based on the mean pixel intensity of the entire dataset.

\subsection{Comparison with 3D Detection Methods}
In this section, we conduct a comprehensive comparison between our proposed method and several state-of-the-art monocular 3D object detection techniques under various adverse weather conditions. These conditions include moderate fog, thick fog, moderate rain, heavy rain, dense rain, and low light. The car category's 3D detection accuracy, denoted by $\mathrm{AP3D}_{R40}$, serves as the benchmark for comparison. The results are presented in Table 1.

Our method demonstrates significant performance improvements across different weather conditions. Under heavy rain conditions, our approach achieves gains of 1.66$\%$, 0.98$\%$, and 0.87$\%$ on the easy, moderate, and hard settings, respectively. Similarly, under thick fog conditions, our method obtains 1.66$\%$, 0.98$\%$, and 0.87$\%$ gains for the same settings. Furthermore, when evaluated on the thick fog dataset, our method outperforms GUPNet by 3.33$\%$, 2.04$\%$, and 1.78$\%$ in terms of 3D detection under the three settings at a 0.7 IoU threshold. Additionally, our MonoTDP method substantially surpasses DID-M3D and MonoDLE in the low light dataset, with improvements of 0.91$\%$ and 3.71$\%$ $\mathrm{AP3D}_{R40}$ under moderate settings. This result serves to validate the effectiveness of our approach.

The superior performance of our method can be attributed to the integration of environmental constraints, as well as the innovative twin depth perception module that concurrently predicts scene depth and object depth. By incorporating both local and global features, our method effectively captures the nuances of various weather conditions and enables more accurate depth estimation. Consequently, our method demonstrates exceptional results across a range of rain, fog, and low light conditions, underscoring its robustness and applicability in diverse real-world scenarios.

\begin{table}[t]
	\centering
	\caption{Comparison of our proposed method with the combinations of our base 3D detection network and popular enhancement models under various challenging conditions for the car category, evaluated using $AP_{R40}$ at IoU threshold of 0.7. We compared our method under thick fog with dehazing models, under heavy rain with deraining models, and under low light with low-light enhancement models, all combined with our base 3D detection model.}
	\vspace{-0.2em}
	\centering
	\renewcommand\arraystretch{0.9} 
	\setlength{\tabcolsep}{2mm}  
	\begin{tabular}{l|c|c|ccc}
	  \toprule[1.25pt]
	  \multirow{2}{*}{Scene} & \multirow{2}{*}{Methods} & \multirow{2}{*}{Venue} & \multicolumn{3}{c}{Car 3D@IOU=0.7} \\
  \cline{4-6}          &       &       & \cellcolor[rgb]{ .906,  .902,  .902}Easy & \cellcolor[rgb]{ .906,  .902,  .902}Mod. & \cellcolor[rgb]{ .906,  .902,  .902}Hard \\
	  \hline
	  \hline
	  \multicolumn{1}{c|}{\multirow{5}{*}{\begin{tabular}[c]{@{}c@{}}Thick \\  Fog\end{tabular}}} & Trans & \textit{CVPR22} & 22.95 & 16.03 & 13.21 \\
			& MSBDN & \textit{CVPR20} & 20.11 & 14.14 & 11.55 \\
			& GCA   & \textit{WACV19} & 21.21 & 14.08 & 12.49 \\
			& DCPDN & \textit{CVPR18} & 19.97 & 13.25 & 11.34 \\
			& Ours  & \textit{-} & \textbf{23.13} & \textbf{16.03} & \textbf{12.95} \\
		\hline
		\hline
	  \multicolumn{1}{c|}{\multirow{5}[2]{*}{\begin{tabular}[c]{@{}c@{}}Heavy \\  Rain\end{tabular}}} & Trans & \textit{CVPR22} & 20.29 & 13.89 & 11.67 \\
			& RESCAN & \textit{ECCV18} & 20.06 & 13.81 & 10.99 \\
			& VRGNet & \textit{CVPR21} & 21.55 & 12.98 & 11.01 \\
			& PRENet & \textit{CVPR19} & 20.11 & 13.34 & 10.67 \\
			& Ours  & \textit{-} & \textbf{23.06} & \textbf{15.77} & \textbf{12.92} \\
		\hline
		\hline
	  \multicolumn{1}{c|}{\multirow{5}[2]{*}{\begin{tabular}[c]{@{}c@{}}Low \\  Light\end{tabular}}} & Trans & \textit{CVPR22} & 14.7  & 10.53 & 9.18 \\
			& SCI   & \textit{CVPR22} & 19.88 & 14.12 & 10.68 \\
			& IAT   & \textit{BMVC22} & 19.84 & 13.59 & 10.94 \\
			& SID   & \textit{CVPR18} & 17.78 & 12.21 & 10.32 \\
			& Ours  & \textit{-} & \textbf{22.55} & \textbf{15.70} & \textbf{12.82} \\
	  \bottomrule[1.25pt]
	  \end{tabular}%
	\label{tab:addlabel}%
\end{table}%

\subsection{Comparison with Restoration Methods}

In this section, we extend our comparison to evaluate the performance of our base 3D detection network combined with various image restoration techniques under different adverse weather conditions. We aim to demonstrate the effectiveness of our adaptive learning strategy against the top-performing defog, rain removal, and low-light enhancement models in thick fog, heavy rain, and low-light environments, respectively. TransWeather is trained in these three weather conditions since it is designed to adapt to various weather scenarios. Other methods are trained under specific environments tailored to their corresponding effects.

As shown in the table 2, the performance comparison between our method and several other methods in dense fog, heavy rain, and low light environments is shown. Among them, the performance of MonoTDP is comparable to TransWeather in dense fog, but shows significant improvement under heavy rain conditions. Under low light conditions it has significantly improved by 7.85\%, 5.17\%, and 3.64\% under the three settings of easy, mod, and hard, respectively. In addition, our method is also significantly superior to all other task specific methods. For example, under dense fog conditions, our method improved by 3.13\%, 2.14\%, and 1.80\% compared to MSBDN under three different settings, respectively. There is also a significant improvement compared to GCA. In heavy rain environments, our method has improved by 1.51\%, 2.79\%, and 1.91\% compared to VRGNet in three settings, and has greater improvement compared to RESCAN. Under low light conditions, in addition to TransWeather, MonoTDP also performs better than SCI and IAT. It can be observed easily from the data that our method has achieved the best performance under various conditions, whether in easy, mod, or hard metrics, which is superior to existing methods. It significantly improves the accuracy and robustness of image restoration in harsh environments.

In summary, our proposed method not only attains state-of-the-art accuracy in harsh environments when compared to the leading 3D object detection techniques, but it also outperforms existing image restoration methods. This result highlights the robustness and adaptability of our approach in diverse and challenging scenarios.

\subsection{Ablation Study}
To investigate how each module in MonoTDP enhances 3D object detection, we randomly selected one seventh of medium rain, heavy rain, dense rain, thin fog, thick fog, dense fog, and low light to obtain a mixed dataset, and then tested each module on this mixed dataset. The results are shown in Table 3.

\begin{table}[t]
	\caption{Ablation study for the components of our method. Results are reported on hybrid datasets contain fog, rain, low light conditions of diverse digrees.}
	\vspace{-0.2em}
	\centering
	\renewcommand{\arraystretch}{0.9}
	\setlength{\tabcolsep}{0.9mm}{
    \begin{tabular}{l||c|c|c|c|ccc}
    \toprule[1.5pt]
    \multirow{2}{*}{} & \multirow{2}{*}{\textbf{WCE}} & \multirow{2}{*}{\textbf{SCD}} & \multirow{2}{*}{\textbf{D$_\mathtt{obj}$}} & \multirow{2}{*}{\textbf{D$_\mathtt{sce}$}} & \multicolumn{3}{c}{3D@IoU=0.7}          \\
    \cline{6-8}              &                                   &                                   &                                       &                                        & \multicolumn{1}{c}{\cellcolor{gray!20} {Easy$\uparrow$}} & \multicolumn{1}{c}{\cellcolor{gray!20} {Mod.$\uparrow$}}    & \multicolumn{1}{c}{\cellcolor{gray!20} {Hard$\uparrow$}}           \\ 
    \midrule[1pt]
    (a)                                   &  -                                 &  -                                 &       -                                &      -                                  & 18.53          & 13.09          & 10.89          \\
    (b)                                   & \checkmark                                 &      -                             &  -                                     &  -                                      & $\mathrm{19.12}_{{\color[HTML]{FF0000}\uparrow 0.59}}$ & $\mathrm{14.43}_{{\color[HTML]{FF0000}\uparrow 1.34}}$ & $\mathrm{11.74}_{{\color[HTML]{FF0000}\uparrow 0.85}}$          \\
    (c)                                   & \checkmark                                 & \checkmark                                 &    -                                   &   -                                     & $\mathrm{20.35}_{{\color[HTML]{FF0000}\uparrow 1.82}}$ & $\mathrm{14.86}_{{\color[HTML]{FF0000}\uparrow 1.77}}$ & $\mathrm{12.11}_{{\color[HTML]{FF0000}\uparrow 1.22}}$          \\ 
    \midrule[1pt]
    (e)                                   &    -                               &     -                              & \checkmark                                     &    -                                    & $\mathrm{20.86}_{{\color[HTML]{FF0000}\uparrow 2.33}}$ & $\mathrm{14.11}_{{\color[HTML]{FF0000}\uparrow 1.02}}$ & $\mathrm{11.86}_{{\color[HTML]{FF0000}\uparrow 0.97}}$          \\
    (f)                                   &     -                              &     -                              & \checkmark                                     & \checkmark                                      & $\mathrm{21.54}_{{\color[HTML]{FF0000}\uparrow 3.01}}$ & $\mathrm{14.25}_{{\color[HTML]{FF0000}\uparrow 1.16}}$ & $\mathrm{12.01}_{{\color[HTML]{FF0000}\uparrow 1.12}}$          \\
    (g)                                   & \checkmark                                & -                                  & \checkmark                                     & \checkmark                                      & $\mathrm{22.18}_{{\color[HTML]{FF0000}\uparrow 3.65}}$ & $\mathrm{14.77}_{{\color[HTML]{FF0000}\uparrow 1.68}}$ & $\mathrm{12.12}_{{\color[HTML]{FF0000}\uparrow 1.23}}$         \\
    (h)                                   & \checkmark                                & \checkmark                                 & \checkmark                                     & \checkmark                                     & $\mathrm{23.22}_{{\color[HTML]{FF0000}\uparrow 4.69}}$ & $\mathrm{15.55}_{{\color[HTML]{FF0000}\uparrow 2.46}}$ & $\mathrm{12.31}_{{\color[HTML]{FF0000}\uparrow 1.42}}$ \\ 
    \bottomrule[1.5pt]
    \end{tabular}
    }
\end{table}

\begin{figure*}[ht]
    \centering
    \includegraphics[scale=0.54]{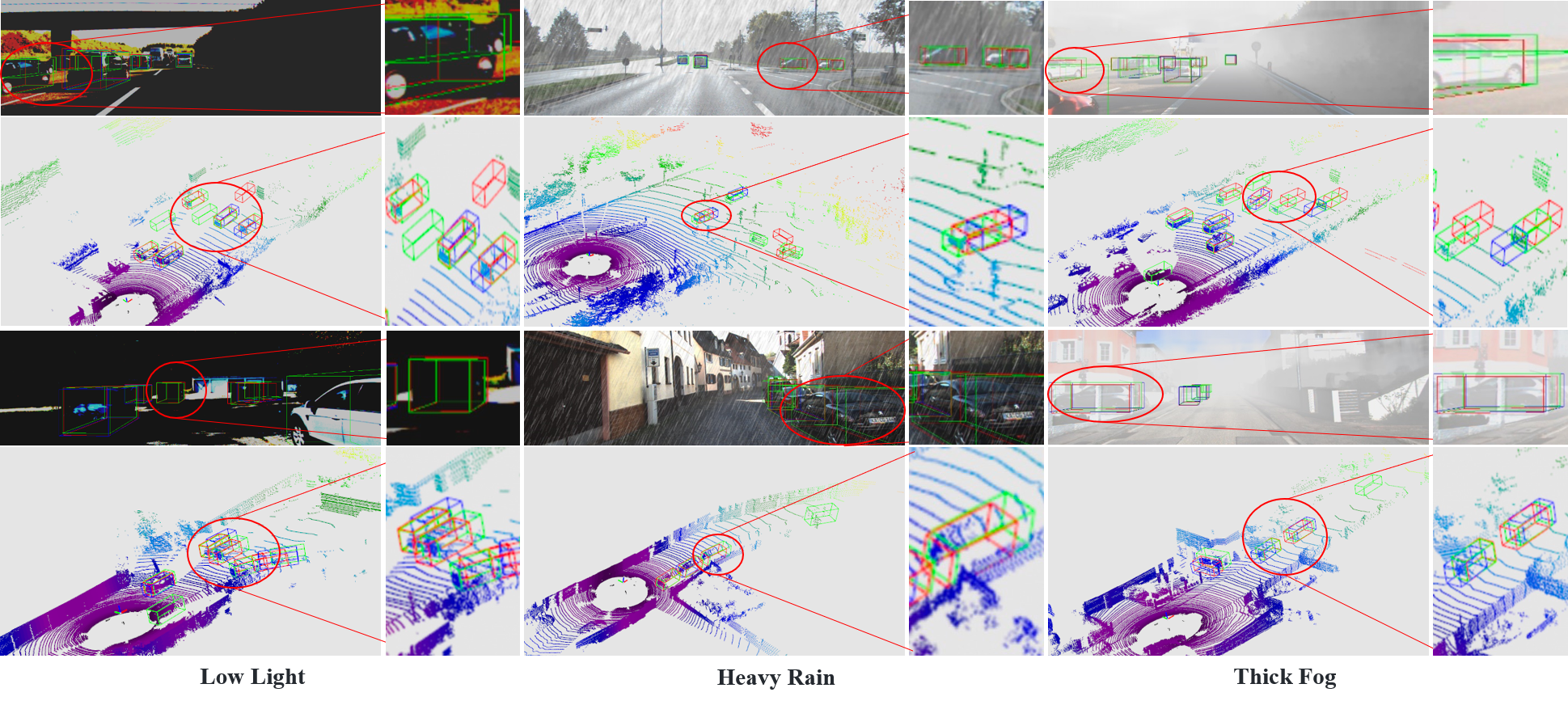}
	\vspace{-1.1em}
	\caption{Qualitative results on the validation set of hybrid dataset which contain all types of weather. These results are based on method trained on the train set. Three columns show the 3D target detection results under different scenes. We use \textcolor{green}{green}, $\,$ \textcolor{red}{red}, $\,$ \textcolor{blue}{blue} boxes to denote ground-truth, our predictions and predictions of GUPNet(One of the most popular 3D Detection Model) respectively. LiDAR signals are only used for visualization.}
\end{figure*}

We evaluate the effectiveness of our adaptive learning strategy which is divided into two parts: Weak Constraint Encoder (WCE) and Strong Constraint Decoder (SCD). We conduct experiments to evaluate the impact of each part on the overall performance of our 3D object detection system.

First, we examine the WCE's contribution by comparing settings (a$\rightarrow$b) and (f$\rightarrow$g). The results demonstrate that the WCE consistently improves the overall performance by 0.34$\%$ for (a$\rightarrow$b) and 0.52$\%$ for (f$\rightarrow$g) under moderate settings. This highlights the WCE's ability to effectively and stably enhance the 3D object detection task.

Next, we assess the SCD's effectiveness through experiments (b$\rightarrow$c) and (g$\rightarrow$h). The observed performance improvement indicates that the SCD is also a crucial component of our learning strategy. Both parts of the module prove to be indispensable for optimal performance.

To further investigate the role of our twin depth perception module, we conduct two sets of control experiments (b$\rightarrow$g and c$\rightarrow$h). The significant improvement observed in these experiments demonstrates that the combination of object depth (D${\mathtt{obj}}$) and scene depth (D${\mathtt{sce}}$) enhances the model's ability to comprehend instance depth.

In particular, the experiments (e$\rightarrow$f and g$\rightarrow$h) reveal that the scene depth design allows the model to obtain more accurate depth estimates across various environments, thereby improving 3D object detection performance. The experimental results conclusively demonstrate the effectiveness of our proposed method.

\subsection{Qualitative Results}
A close examination of the information presented in Figure 5 reveals the superior performance of our proposed method, MonoTDP, compared to the current state-of-the-art approach, GUPNet, in three distinct environments: low light, rainy, and foggy conditions. In low light scenarios, where the environment is comparatively dark, GUPNet tends to miss objects, whereas MonoTDP demonstrates higher accuracy in recognizing the majority of images. Under rainy conditions, GUPNet struggles to correctly identify numerous objects due to rain-induced obstructions. In contrast, our method exhibits minimal target misidentification, thereby addressing a significant limitation of GUPNet. In foggy situations characterized by low visibility, GUPNet's recognition results often deviate considerably from the correct outcomes, leading to inaccurate object detection. Conversely, MonoTDP accurately identifies nearly all objects. These observations highlight the considerable advantages of our method over other optimal approaches, emphasizing its resilience against environmental challenges. MonoTDP effectively tackles issues, such as texture loss, rain streak occlusions, and impaired visibility, which are prevalent in real-world 3D object detection tasks.

\section{Conclusion}
In this study, we introduce MonoTDP, a monocular 3D object detection model adept at perceiving twin depth and demonstrating exceptional performance in an array of challenging environments, including fog, rain, and low-light conditions. Incorporating an adaptive learning strategy and a twin depth perception module, our model enhances the accuracy of 3D object detection, even under adverse circumstances.
MonoTDP effectively utilizes the adaptive learning strategy to regularize the model, enabling it to adapt to inclement weather conditions and perceive features across diverse scenes. Simultaneously, the model estimates both scene depth and object depth, thus rendering the depth prediction process scene-aware. This innovative approach substantially advances the practical applicability of monocular 3D object detection models.
Extensive experimental results attest to the superiority of our proposed method over state-of-the-art approaches, both qualitatively and quantitatively, across various adverse environments.

\newpage
\balance
\bibliographystyle{ACM-Reference-Format}
\bibliography{egbib}
\end{document}